\documentclass{article}

\PassOptionsToPackage{numbers, compress}{natbib}

\usepackage[preprint]{neurips_2020}

\usepackage[utf8]{inputenc} 
\usepackage[T1]{fontenc}    
\usepackage{hyperref}       
\usepackage{url}            
\usepackage{booktabs}       
\usepackage{nicefrac}       
\usepackage{microtype}      
\usepackage{graphicx}

\usepackage{tabularx}
\usepackage{adjustbox}

\usepackage{subcaption}
\usepackage{amsmath,amsfonts,amssymb}
\usepackage{xcolor}
\usepackage[ruled,vlined]{algorithm2e}
\usepackage{floatrow}

\usepackage{colortbl}

\title{Tuning a variational autoencoder for data accountability problem in the Mars Science Laboratory ground data system}

\author{%
  Dounia Lakhmiri\\
  GERAD and \\
  Polytechnique Montréal\\
  \texttt{dounia.lakhmiri@gerad.ca} \\
  \And
  Ryan Alimo \\
  Jet Propulsion Laboratory \\
  California Institute of Technology
  \\
  \texttt{sralimo@jpl.nasa.gov} \\
  \And
  Sébastien Le Digabel \\
  GERAD and \\
  Polytechnique Montréal\\
  \texttt{sebastien.le.digabel@gerad.ca} \\
}

\begin{document}

\maketitle

\begin{abstract}
  The Mars Curiosity rover is frequently sending back engineering and science data that goes through a pipeline of systems before reaching its final destination at the mission operations center making it prone to volume loss and data corruption. A ground data system analysis (GDSA) team is charged with the monitoring of this flow of information and the detection of anomalies in that data in order to request a re-transmission when necessary. This work presents $\Delta$-MADS, a derivative-free optimization method applied for tuning the architecture and hyperparameters of a variational autoencoder trained to detect the data with missing patches in order to assist the GDSA team in their mission.
\end{abstract}

\textbf{Keywords:} Anomaly detection, variational Autoencoder, hyperparameter optimization, architecture search, derivative-free optimization.

\section{Introduction}
In the NASA Mars Science Laboratory (MSL), a ground data system analysis (GDSA) team is tasked with the analysis of telemetry data sent by the Mars Curiosity rover that travels through a pipeline of satellites and receptors. During its journey back to Earth, this data can be subjected to corruption and volume loss that needs to be detected efficiently in order to ask for a re-transmission when necessary. This problem is akin to an anomaly detection task where one must learn from unlabelled data to differentiate between a normal behaviour and outliers in order to identify the anomalous data. This task is so far handled manually by human experts and needs to be automated to speed up the treatment process and possibly increase the detection accuracy.

In the recent years, deep learning algorithms have shown their efficiency in solving several challenging regression and classification problems~\cite{balakumar2020machine, deng2018deep, jiao2019survey} thanks in parts to the ever growing performances of deep neural networks. These computational graphs can learn from complex, real world datasets and make predictions with an accuracy that can sometimes surpass human experts. This study focuses on a particular type of autoencoders (AE) which are deep neural networks (DNNs) used for data compression, matrix factorization, anomaly detection, etc. AEs are unsupervised or semi-supervised learning algorithms that start by compressing the input data to map it into a lower dimension latent space before decompressing it back to recreate the original input. The learning phase aims at recreating the input data as closely as possible by minimizing the reconstruction error between the original data and its reconstruction as represented in Figure~\ref{fig:ae_arch}. The usual framework for anomaly detection with AEs is to collect the reconstruction errors for all points of the dataset and find a threshold value that will determine which errors are considered outliers. AEs have proven themselves to be efficient tools for unsupervised anomaly detection~\cite{dlamini2019lightweight, provotar2019unsupervised} with the caveat that the architecture and hyperparameters of such neural networks must be adequately chosen to get a competitive performance for real life applications.  

As for any neural network, the choices for the hyperparameters that define the architecture as well as the training phase have a great impact on the overall precision of the network and its ability to generalize. This tedious and consuming process, in terms of time and computational power, can be modeled as a blackbox optimization problem. Blackbox optimization is a subfield of derivation-free optimization (DFO)~\cite{AuHa2017, CoScVibook}, a discipline that considers optimization problems without relying on derivatives since they may not exist or are too complex to compute. DFO also covers the case where a function evaluation is the result of an expensive computation or a simulation, seen as blackboxes, that can fail at some points. In this case, the blackbox is defined so that its input is a particular configuration and the output is the test error of the corresponding network after it is trained on the data set. This work presents a new approach named $\Delta$-MADS obtained by merging two DFO schemes, HyperNOMAD~\cite{hypernomad_paper} and $\Delta$-DOGS~\cite{beyhaghi2016delaunay} in order to exploit the strong suits of each. The resulting algorithm is applied on the previously described tuning problem.

The remaining of the paper is organized as follows: Section~\ref{sec:litterature} gives an overview of anomaly detection techniques and of the main approaches used to solve the HPO problem of deep neural networks. Section~\ref{sec:vae} presents Variational Autoencoders, their architecture, how they are trained for anomaly detection problems and the hyperparameters focused on in this paper. Section~\ref{sec:hybrid} describes the hybrid DFO algorithm which is tested on this particular problem and benchmarked against other optimization schemes in Section~\ref{sec:experiment}. Finally, Section~\ref{sec:conclusion} synthesizes the results in a short conclusion.

\section{Related work}
\label{sec:litterature}

Anomaly detection is an expanding field of research with many real life applications such as credit card fraud detection~\cite{carcillo2019combining}, finding network intrusions in the context of cyber security~\cite{vartouni2018anomaly}, industrial damage detection~\cite{yan2019accurate}, etc. These problems usually amount to a classification task on imbalanced data, meaning that the outliers represent more often than not a small fraction of the overall data set. Also, depending on the data set, this problem can be a supervised, a semi-supervised or an unsupervised task. This work focuses on the two latter cases since the labels of the training data are the only ones available. Some popular clustering methods such as Gaussian mixtures, K-Means, DBSCAN, etc. can be applied on unsupervised anomaly detection problems~\cite{agrawal2015survey, emadi2018novel, li2016anomaly} but they often fall short when dealing with high dimensional data with complex structures contrary to deep learning algorithms. In a semi-supervised or unsupervised context, generative neural networks such as AEs can be adapted for anomaly detection problems~\cite{dlamini2019lightweight, provotar2019unsupervised, schreyer2017detection} by training them on normal data so that they learn to reproduce or generate good behaviors with a small reconstruction error. During this training, each data point is first reduced to a lower dimension representation that is expanded to its original size afterwards. The implicit hypothesis is that outliers should be different enough from normal data so that a trained AE will get a higher reconstruction error on outliers than on a data that behaves like the normal training points. However, some real life applications do not satisfy this hypothesis such as the one considered in this study. Indeed, all the data received by the MSL goes through the same steps and systems and has therefore the same underlying structure which represents a challenge for AEs that struggle with separating the normal behavior from the anomalies.  

Variational autoencoders (VAEs)~\cite{kingma2013autoencoding} are generative, probabilistic graphical models that share a similar architecture with regular AEs as shown in Figure~\ref{fig:vae_arch} with the main difference residing in the latent representation of the data. Instead of mapping each input point to a deterministic lower dimensional vector, a VAE maps it with a region in the latent space by learning the parameters of a probability distribution that approximates the posterior. As shown in~\cite{an2015variational}, VAEs can surpass normal AEs in anomaly detection problem such as finding the outliers on the MNIST data set. Further details on VAEs and their training are presented in Section~\ref{sec:vae}.

As any DNN, VAEs are extremely sensitive to their structure, or architecture, and to the values of the hyperparameters related to the optimization process that happens during the training phase. Many different approaches were explored to automate the search for optimal hyperparameters starting with the grid search which evaluates all the possible combinations of hyperparameters from a constrained search space. This method is clearly expensive and does not scale well with the dimension of the problem. The random search~\cite{bergstra2012random} has been shown to be more effective than grid search but is still highly expensive and lacks adaptiveness to the problem. Bayesian methods offer a more sophisticated alternative by either constructing a model over the objective function $f$ in the case of a Gaussian processes~\cite{williams2006gaussian} or random forests~\cite{breiman2001random}, or over the distribution of the good and bad configurations in the case of tree parzen estimators~\cite{bergstra2011algorithms}, by using the previously evaluated points. Other approaches were tested such as reinforcement learning~\cite{baker2016designing, zoph2016neural} which is successfully  used to find the appropriate architecture of convolutional neural networks, and more recently the HyperNOMAD~\cite{hypernomad, hypernomad_paper} software, based on the mesh adaptive direct search (MADS) algorithm~\cite{audet2006mesh}, was able to yield good results when optimizing both the architecture and the training hyperparameters simultaneously. The main drawback of this software is its lack of global exploration strategy. A hybrid algorithm is proposed in this work that combines the local refinement of HyperNOMAD with the global search of $\Delta$-DOGS~\cite{beyhaghi2016delaunay}, another DFO algorithm equipped with a global search model based on Delaunay's triangulation which was shown to explore efficiently the search space on smaller dimension problems and on limited types of variables. This new approach manages to exploit the advantages of each method and is further discussed in Section~\ref{sec:hybrid}. 

\section{Anomaly Detection with Variational Autoencoders}

\label{sec:vae}
The following section provides a high-level description of variational autoencoders (VAEs), their architecture and training before going through the list of hyperparameters considered for the anomaly detection problem.

\subsection{Overview of variational autoencoders}

\begin{figure}[h]
    \centering
    \hspace{-1cm}\begin{subfigure}[t]{0.45\textwidth}
        \centering
        \includegraphics[scale=0.27]{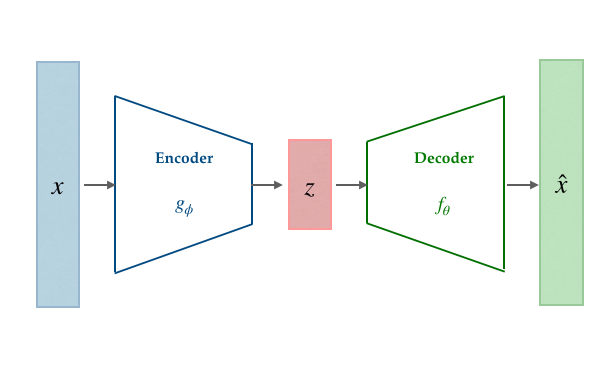}
            \caption{Autoencoder architecture: the input vector $x$ is passed to the encoder $g_\phi$ that produces a lower dimension representation $z$ which is fed to the decoder $f_\theta$ to produce the reconstruction $\hat{x}$.}
            \label{fig:ae_arch}
    \end{subfigure} \hspace{1cm}%
    \begin{subfigure}[t]{0.4\textwidth}
        \centering
        \includegraphics[scale=0.27]{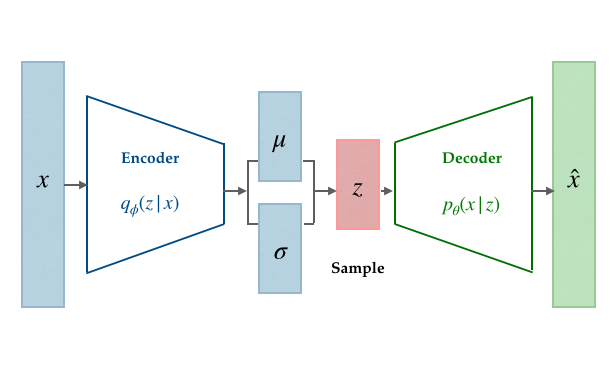}
        \caption{Variational autoencoder architecture: the input vector $x$ is passed to the encoder $q_\phi(z|x)$ that produces the mean $\mu$ and standard deviation $\sigma$ of a normal distribution from which a sample $z$ is drawn to be fed to the decoder $p_\theta(x|z)$ to produce the reconstruction $\hat{x}$.}
        \label{fig:vae_arch}
    \end{subfigure}
    \caption{Structure of an autoencoder on the left and of a variational autoencoder on the right.}
\end{figure}

A VAE, as shown in Figure~\ref{fig:vae_arch}, is a deep neural network made up of three sections: an encoder defined by the weights $\phi$, an encoding layer in the middle of the network of size $n_e$ and a decoder defined by the weights $\theta$. Let $x \in \mathbb{R}^{n_0}$ be an input vector which is first passed to the encoder that reduces its dimension from layer to layer until reaching the middle of the network which has the smallest size: $n_e$. At this point, the VAE generates two vectors $\mu, \sigma \in \mathbb{R}^{n_e}$ that represent the mean and variance of a normal distribution from which a sample $z \in \mathbb{R}^{n_e}$ is drawn. Therefore, the VAE associates the input vector $x$, and consequently its class, with a region in the latent space where its lower dimensional representation $z$ is more likely to be. Note that the choice of the normal distribution is used for practical purposes and can be altered if needed. The role of the encoder of a VAE is changed from a deterministic compressing function in the case of a standard AE, to a probabilistic model that learns the distribution of the latent representation $z$ for a given input $x$ noted $q_\phi(z|x)$.
The second phase consists of passing the latent vector $z$ to the decoder that expands it from layer to layer until forming a reconstruction $\hat{x} \in \mathbb{R}^{n_0}$ which is compared to the original input $x$. Once again, the decoder is no longer the deterministic function of a standard AE, but is now a probabilistic model $p_\theta(x|z)$ that learns the distribution of $\hat{x}$, and therefore of $x$, knowing the input $z$.

Therefore, the latent representation $z$, and consequently $\hat{x}$ are not deterministic for the same input $x$ which poses a challenge during the training of the VAE, especially when the error is backpropagated through the network to update its weights. The reparametrization trick~\cite{kingma2013autoencoding} was introduced to solve this exact problem: the latent representation is now calculated as $z = \mu + \epsilon \sigma$ with $\epsilon \sim N(0, 1)$ instead of randomly sampling $z \sim N(\mu, \sigma)$ directly. By moving the random sampling to $\epsilon$, the backpropagation, which can not be applied on a stochastic term, can effectively reach the layers $\mu, \sigma$ and the rest of the encoder.

The loss function of the VAE is composed of a term that quantifies the reconstruction error between the input $x$ and the reconstruction $\hat{x}$, plus a regularization term on the latent space to ensure that the encoded distribution is close to a standard normal distribution using the Kulback-Leibler divergence as seen in the following Equation:
\begin{equation}
    \label{eq:loss_vae}
    L = ||\hat{x} - x|| + D_{KL}(q_\phi(z|x) || p(z))
\end{equation}
where $p(z) \sim N(0, I)$ and $D_{KL}$ is the  Kulback-Leibler divergence term between $q_\phi(z|x)$ and $p(z)$.
This regularization gives desirable properties to the latent space by ensuring a good distribution of the latent variables which is essential for a generative model~\cite{asperti2018sparsity}.

\subsection{Hyperparameters of a VAE}

This study focuses on tuning both the architecture and the learning hyperparameters to obtain an effective VAE for a particular anomaly detection task. The number of variables defining the architecture can be significantly reduced by considering a symmetric VAE with layers of decreasing, respectively increasing, size in the encoder, respectively decoder. The architecture can therefore be described by two integer variables, one representing the number of encoding layers and the second for the dimension of the latent space. The size of the remaining encoding, respectively decoding, layers can be deduced from this two values by imposing a linear decrease, respectively increase.  The activation function is used to introduce a nonlinear transformation after each layer of the encoder, respectively the decoder. The dropout rate is added as a regularization technique to avoid the over-fitting issue. As for the training phase, the batch size determines the number of training data points passed to the VAE at the same time which affects both the learning of the network and the speed of the training. Additionally, the choice of the optimizer algorithm along with four hyperparameters are also considered in this tuning problem which are summarized in Table~\ref{tab:optimizer}. 

\begin{table}[h]
    \caption{Hyperparameters related to the training of the VAE.}
    \label{tab:optimizer}
    \centering
    \begin{tabular}{p{0.4 \textwidth}  p{0.25 \textwidth}  p{0.1 \textwidth}  p{0.1 \textwidth}} \toprule
    	\textbf{Optimizer } & \textbf{Hyperparameter} & \textbf{Type} & \textbf{Range} \\ \midrule
	Stochastic Gradient Descent (SGD)	& Initial learning rate 		& Real 	& [0;1]\\
	 						       	& Momentum 			& Real 	& [0;1]\\
							       	& Damping 			& Real 	& [0;1]\\
							       	& Weight decay 		& Real 	& [0;1]\\ \midrule
	Adam						& Initial learning rate 		& Real 	& [0;1]\\
								& $\beta_1$ 		        & Real 	& [0;1]\\
								& $\beta_2$			        & Real 	& [0;1]\\
								& Weight decay 		        & Real 	& [0;1]\\ \midrule
	Adagrad						& Initial learning rate 	& Real 	& [0;1]\\
								& Learning rate decay	    & Real 	& [0;1]\\
								& Initial accumulator		& Real 	& [0;1]\\
								& Weight decay 		        & Real 	& [0;1]\\ \midrule
	RMSProp						& Initial learning rate 	& Real 	& [0;1]\\
								& Momentum 			        & Real 	& [0;1]\\
								& Smoothing constant	    & Real 	& [0;1]\\   
								& Weight decay 		        & Real 	& [0;1]\\ 		       
	 \bottomrule
   \end{tabular}
\end{table}

where $\beta_1$ is the factor for the first moment estimates and $\beta_2$ is factor for the second moment estimates.

In practice, VAEs can be used for a semi-supervised anomaly detection problem by applying the following protocol: the network is trained on normal data so that it learns to replicate normal behaviors only which results in bigger reconstruction errors when anomalous data is passed through the VAE. In order to classify each input data, the generated error is compared to a certain threshold value $\alpha \in \mathbb{R}$ which can either be fixed by the user according to some acquired knowledge on the classification task at hand, or has to be also tuned as an additional hyperparameter which is the case for the application considered here.

\begin{table}[h]
    \centering
    \caption{List of the hyperparameters of the VAE considered for the tuning problem.}
    \begin{tabular}{p{0.5 \textwidth}  p{0.2 \textwidth}  p{0.2 \textwidth}} \toprule
   	\textbf{Hyperparameter } & \textbf{Type} & \textbf{Range} \\ \midrule

        Number of encoding layers & Integer & $[1, 50]$ \\ \midrule
        Dimension of the latent space & Integer & $[1, n_0[$ \\ \midrule
        Batch size          & Integer       & $[10, 512]$ \\ \midrule
                            &               & $1:$ ReLU, \\
        Activation function & Categorical   & $2:$ Sigmoid,\\
                            &               & $3:$ Tanh. \\ \midrule
        Dropout rate        & Real          & $[0, 1]$ \\ \midrule
        Optimizer choice    & Categorical   & $1:$ SGD, \\
                            &               & $2:$ Adam. \\
                            &               & $3:$ Adagrad. \\ 
                            &               & $4:$ RMSProp. \\ \midrule
        $4$ HPs of the optimizer (Table~\ref{tab:optimizer}) & Real     & $[0, 1]$ \\ \midrule
        Threshold $\alpha$  & Real          & $[0.50, 1]$ \\
       \bottomrule
    \end{tabular}
    
    \label{tab:hpo}
\end{table}

\section {The $\Delta$-MADS method}
\label{sec:hybrid}
This section describes $\Delta$-MADS, a hybrid algorithm that mixes the local search of HyperNOMAD~\cite{hypernomad, hypernomad_paper} with the global exploration scheme of $\Delta$-DOGS~\cite{beyhaghi2016delaunay}. $\Delta$-MADS is designed to solve derivative-free optimization problems formulated as follows:
\begin{equation}\label{eq:objective_function}
    \min_{x \in \Omega} f(x)
\end{equation}
where $\Omega = \{x \in \mathbb{R}^n \quad | \quad a \leq x \leq b  \text{ with } a, b \in \mathbb{R}^n\}$. The notation $x^N$ refers to the integer and categorical components of the vector $x$ and $x^R$ the real elements of $x$. The entire vector $x$ can be reconstructed by combining $x^N$ and $x^R$ which is written as $x = x^N \cup x^R$. In the context of this specific application, tuning a variational autoencoder can be modeled as a derivative-free, and more specifically a blackbox, optimization problem where the objective function $f$ takes a set of hyperparameters, builds the corresponding variational autoencoder, trains, validates and tests its performance before returning the mean $F1$ score, described in Section~\ref{sec:experiment}, on the test set as a measure of performance.

HyperNOMAD, being based on MADS~\cite{audet2006mesh}, is an iterative algorithm with two phases. The first one, called the \textit{search}, is an optional and flexible step where a global optimization scheme can be implemented and the second one, called the \textit{poll}, is rigorously established. At each iteration $k$, the mesh $M_k = \{x + \Delta_k^{m}Dz, z \in \mathbb{N}^{n_D}, x \in C\}$ is defined where $C$, called the cache, is the list that stores all of the previously evaluated points, the matrix $D \in \mathbb{R}^{n \times n_D}$ has columns that form a positive spanning set and $\Delta_k^{m} \in \mathbb{R}^{+}$ is the mesh size. The \textit{poll} starts around the current point $x_k$ and defines the poll set $P_k = \{x_k + \Delta_k^{m}d \mid d \in D_k\} \text{ with } ||\Delta_k^{m}d|| \approx \Delta_k^{p}$, that contains the candidates evaluated opportunistically, meaning that the \textit{poll} step will end as soon as a better point is found. In that case, a new iteration starts with a larger mesh size and otherwise, the mesh size is reduced. HyperNOMAD is adapted to handle real, integer and categorical variables\cite{AACW09a, AuLeDTr2018}. The later requires to define an \textit{extended poll}~\cite{AuLeDTr2018} which links the different search spaces related to different values of the categorical variables. The MADS algorithm offers a hierarchy of convergence results depending on the properties of the optimization problem. In~\cite{audet2006mesh}, the authors prove that MADS converges to a 
Clarke, respectively contingent KKT, stationary point if $f$ is Lipschitz near the limit point, respectively strictly differentiable at the limit point. The convergence of MADS is derived solely from the \textit{poll} step and is maintained if the \textit{search} generates a finite number of trial points each time it is called which are then projected onto the mesh $M_k$ at iteration $k$.

$\Delta$-DOGS is a family of iterative derivative-free optimization methods~\cite{alimo2020delaunay, alimo2020design, beyhaghi2016delaunay} that rely on a surrogate model of the objective function to direct the optimization. The surrogate search function $s$ is computed at each iteration by combining an interpolation function $p$ with an artificially generated uncertainty function $e$ based on Delaunay's triangulation that plays a similar role to the acquisition functions in a Bayesian optimization scheme so that $s(x) = p(x) - K e(x), x \in \mathbb{R}^n$. The tuning parameter $K$ depends on the target value $y^*$ that the user hopes to reach during the optimization. $\Delta$-DOGS is proven to globally converge in the case of convex optimization problem where the Lipschitz bound of the objective function is bounded~\cite{beyhaghi2016delaunay}, however it scales poorly to the dimension of the optimization problem $n$ and is not adapted to handle mixed variable problems. 

The $\Delta$-MADS algorithm~\ref{algo:hybrid} mixes aspects of the two DFO schemes previously described by implementing the surrogate function of $\Delta$-DOGS into the \textit{search} step of HyperNOMAD. Also, while the \textit{poll} step is optimizing the entire set of hyperparameters listed in Section~\ref{sec:vae}, the \textit{search} phase keeps the integer and categorical variables $x_k^N$ fixed and optimizes the sub-problem considering only the continuous variables $x_k^R$ therefore reducing the dimension of the problem and interpolating only on continuous functions. The algorithm starts with an initial point $x_0$ and a target value $y_0$ passed onto the search of $\Delta$-DOGS which is given a certain budget of function evaluations, the best solution found in the \textit{search} is passed to the \textit{poll} from which the local refinement starts. The target value $y_k$ is re-evaluated at each iteration depending on whereas it was achieved or not. For this minimization problem, the target $y_k$ is decreased if $y_{k-1}$ was attained or improved upon, and increased if not. The algorithm alternates this way between the two phases until the function evaluation budget is depleted or a convergence condition is achieved. The convergence properties of this novel approach are inherited from MADS since the \textit{search} step is guaranteed to produce a finite number of mesh candidates.
\begin{algorithm}[h]
    \SetAlgoLined
        initialization: $x_0, y_0, \epsilon \in ]0, 1[, k=0$ \;
        \While{not stop}{
            \vspace{2mm} \textbf{Search step: } Fix the integer and categorical variables and apply $\Delta$-DOGS on the remaining variables $x_k^R$ with the target $y_k$ \; 
            
            return new ${x^{'}}_k^R$, reconstruct the complete vector $x_k^{'} = x_k^N \cup {x^{'}}_k^R$ and project it on the mesh \; 
            
            \vspace{2mm} \textbf{Poll step: } Apply HyperNOMAD on the new $x_k^{'}$ and return the best feasible solution found $x_{k+1}$ \;

            \vspace{2mm} \textbf{Updates: }
            Set $f_{k+1}$ the best objective value \;
            
            \eIf{$f_k < y_k$}{
                $y_{k+1} = y_k - \epsilon$\;
            }{
                $y_{k+1} = y_k + \epsilon$\;
            }
        }
    \caption{$\Delta$-MADS : Hybrid between HyperNOMAD and $\Delta$-DOGS}
    \label{algo:hybrid}
\end{algorithm}

\vspace*{-5mm}\section{Numerical results for the MSL data accountability}
\label{sec:experiment}

The Mars Curiosity rover transmits telemetry data to the MSL ground system operations team through a complex pipeline of systems where each transfer leaves the data inevitably susceptible to corruptions.
In order to increase the traceability in this process, the ground data system (GDS) records metadata about the transmissions at three locations in the downlink process: the orbiter used to transmit the data, JPL Data Control, and the data’s final destination in the MSL GDS. 
This metadata is analyzed by experts to determine if the original data is successfully transferred, called a complete pass, or not, called an incomplete pass.

This work considers the latest dataset available at this time which includes a total of $9805$ passes, $8493$ of which are complete passes and the remaining $1312$ are incomplete. With this proportion of around $13\%$ anomalies, a $87\%$ classification accuracy can easily be achieved simply by labelling every pass as a complete one which shows the limitation of usual accuracy metrics when dealing with unbalanced datasets. In this case, other metrics such as the precision $P$, recall $R$ and $F1$ score, shown in Equation~\ref{eq:metrics}, are more adequate to correctly evaluate the quality of a classifier. For each class, the precision $P$ measures how many of the classifier's predicted labels are correct. The recall $R$ quantifies how many true members of a certain class are correctly identified and the $F1$ score combines both metrics as follows:\begin{equation} \label{eq:metrics}
    P = \frac{TP}{TP + FP}, \quad
    R = \frac{TP}{TP + FN}, \quad
    F1 = 2 \frac{R \times P}{R + P}
\end{equation}
where $TP$ is the number of true positives, $TN$ the number of true negatives and $FN$ the number of false negatives. 

As an initial step toward solving this anomaly detection problem, different unsupervised methods are implemented without hyperparameter optimization. The results of these different methods are compiled in Table~\ref{tab:comp_unsupervised} and compared against the GDS labeler, which is the previous algorithm used by the GDSA team to identify incomplete passes. The GDS Labeler yields the best results for correctly identifying complete passes, however the recall of $55\%$ on incomplete passes means that only $55\%$ of the anomalies are correctly detected as such. Plus, the GDS Labeler takes about $5$ hours to complete the anomaly detection task. The VAE, whose hyperparameters are chosen without any prior intuition, gets the best results on correctly detecting incomplete passes compared to the rest of the detection methods which justifies the hyperparameter optimization effort conducted especially considering that its training, validation and testing takes around $50$ seconds.

\begin{table}
\caption{Performance of the unsupervised learning methods: KMEAN, Gaussian mixtures, autoencoder and variational autoencoder without hyperparameter optimization on the missing data detection problem compared against the current GDS labeler.}
    \centering
    \begin{tabular}{lcccccccccl}\toprule
 & \multicolumn{2}{p{0.15 \textwidth}}{GDS labeler} & \multicolumn{2}{p{0.1 \textwidth}}{KMEAN} & \multicolumn{2}{p{0.15 \textwidth}}{Gaussian mixture} & \multicolumn{2}{p{0.1 \textwidth}}{Untuned AE} & \multicolumn{2}{p{0.1 \textwidth}}{Untuned VAE} 
            \\\cmidrule(lr){2-3}\cmidrule(lr){4-5} \cmidrule(lr){6-7} \cmidrule(lr){8-9} \cmidrule(lr){10-11} 
           & Cpl.   & Inc.     &  Cpl. & Inc.     & Cpl. & Inc. & Cpl. & Inc. & Cpl. & Inc. \\ \midrule
$P$  & $0.94$  & $0.74$   & $0.08$    & $0.40$    & $0.26$   & $0.30$ & $0.55$ & $0.74$  & $0.84$ & $0.75$ \\
$R$      & $0.97$  & $0.55$   & $0.01$    & $0.86$   & $0.15$   & $0.46$ & $0.63$ & $0.62$ & $0.79$ & $0.80$ \\
$F1$    & $\mathbf{0.95}$  & $0.63$   & $0.02$    & $0.55$   & $0.19$   & $0.36$ & $0.58$ & $0.67$ & $0.81$ & $\mathbf{0.77}$ \\ \bottomrule
\end{tabular}
    \label{tab:comp_unsupervised}
\end{table}
The blackbox that models this problem takes all the hyperparameters listed in Section~\ref{sec:vae} as inputs, constructs the corresponding VAE and splits the data into three sets: training, validation and test with the training set containing complete passes only and the remaining two are a mix of normal behavior and anomalies. After the training and validation of the VAE, the blackbox returns the average of the $F1$ scores on complete and incomplete passes on the test set as a performance measure. The comparison is conducted with two different starting points, one that gives an initial $F1$ score of $81\%$, refereed to as a advantageous initialization, and the other gives an initial $F1$ score of $65\%$ which is refereed to as the disadvantageous initialization. The goal here is to observe the behavior of the HPO algorithms in two different settings.

\begin{figure}[h]
    \centering
    \hspace{-1cm}\begin{subfigure}[t]{0.45\textwidth}
        \centering
        \includegraphics[scale=0.4]{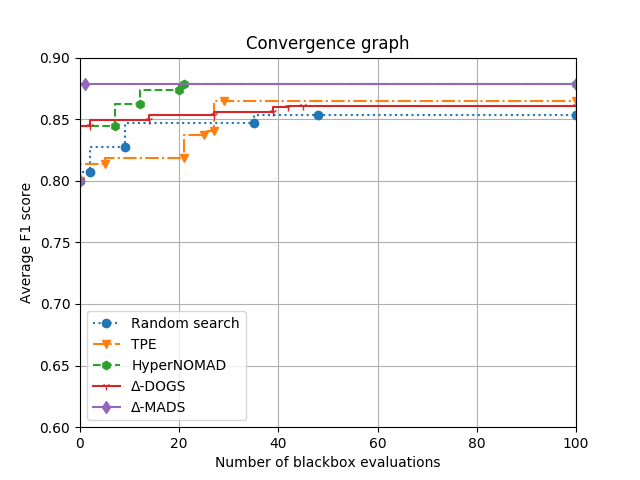}
            \caption{Convergence graphs for each HPO algorithm on the advantageous initialization.}
            \label{fig:benchmark}
    \end{subfigure} \hspace{0.5cm}%
    \begin{subfigure}[t]{0.4\textwidth}
        \centering
        \includegraphics[scale=0.4]{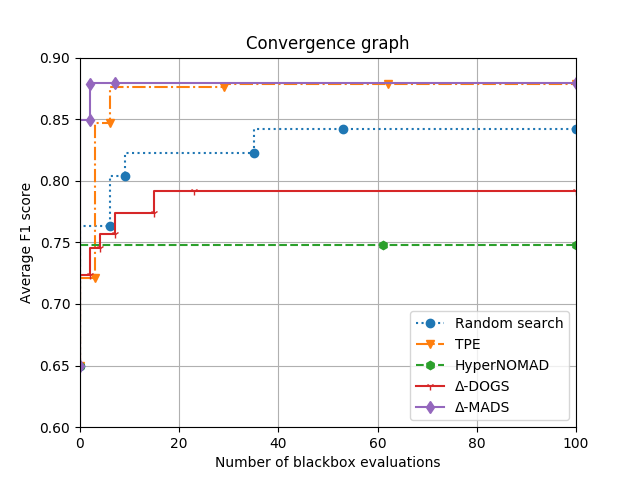}
        \caption{Convergence graphs for each HPO algorithm on the disadvantageous initialization.}
        \label{fig:benchmark_bad}
    \end{subfigure}
    \caption{Comparison between $5$ hyperparameter optimization algorithms: random search, tree Parzen estimator, HyperNOMAD, $\Delta$-DOGS and $\Delta$-MADS, through their convergence curves on two different initializations.}
\end{figure}

Table~\ref{tab:benchmark} compiles the scores of the best configuration found by each hyperparameter optimization method: random search (RS), tree parzen estimator (TPE), HyperNOMAD, $\Delta$-DOGS and $\Delta$-MADS presented in Section~\ref{sec:hybrid}, starting from the advantageous, respectively the disadvantageous, initialization. Both the random search and the tree parzen estimator are tested through the Hyperopt library~\cite{bergstra2013making}. Each solution is evaluated, in the sense of the blackbox, five times and the mean and standard deviation of each score are reported in Table~\ref{tab:benchmark}. In both cases, the results show that all the hyperparameter optimization schemes were able to improve on the original VAEs, thus proving the necessity of this tuning effort in a real life application. 
Starting with the advantageous initialization, the scores of the best solutions obtained by each scheme can be split into two sets: the random search, tree Parzen estimator and $\Delta$-DOGS ended with configurations with an $F1$ score of $85-86\%$ on complete passes and $84\%$ on incomplete passes and both HyperNOMAD and the $\Delta$-MADS obtained the best solutions with an $F1$ score of $88\%$ on complete passes and $87\%$ on the incomplete ones. The advantage of $\Delta$-MADS is better highlighted through Figure~\ref{fig:benchmark} that shows how the $\Delta$-MADS algorithm is always ahead of HyperNOMAD for a certain budget of blackbox evaluations. The hybrid algorithm allows then to reduce the computational cost of the HPO problem. The example with the disadvantageous configuration gives more contrasted results: HyperNOMAD gets the worst results on both complete and incomplete passes which highlights its difficulty to move from a disadvantageous starting point. $\Delta$-DOGS has a similar $F1$ score of $76\%$ on the complete passes and a better performance on the incomplete ones with an $F1$ score of $79\%$ which is believed to be a consequence of the presence of integer and categorical variables in the HPO problem. The best $F1$ score of $87\%$ on the complete passes is obtained by both $\Delta$-MADS and the tree parzen estimator and $\Delta$-MADS achieves the best results of  $87\%$ on the incomplete ones. Once again, $\Delta$-MADS  does so with significantly less blackbox evaluations than any other HPO scheme as is shown in Figure~\ref{fig:benchmark_bad}.
All the HPO methods require an execution time between $1,5$ and $2$ hours to evaluation $100$ configurations with each blackbox evaluation lasting from $35$ to $70$ seconds.

\begin{table}
	\centering
	\caption{Comparison between the scores of the best VAE found by each hyperparameter optimization method with the advantageous initialization (top) and the disadvantageous one (bottom).}
     \label{tab:benchmark}
    \begin{adjustbox}{width=\textwidth}
    	\begin{tabular}{lcccccccccl}\toprule
 & \multicolumn{2}{p{0.15 \textwidth}}{RS} & \multicolumn{2}{p{0.15 \textwidth}}{TPE} & \multicolumn{2}{p{0.15 \textwidth}}{HyperNOMAD} & \multicolumn{2}{p{0.12 \textwidth}}{$\Delta$-DOGS} & \multicolumn{2}{p{0.12 \textwidth}}{$\Delta$-MADS}   
            \\\cmidrule(lr){2-3}\cmidrule(lr){4-5} \cmidrule(lr){6-7} \cmidrule(lr){8-9} \cmidrule(lr){10-11}
            & Cpl.   & Inc.   & Cpl.   & Inc.   & Cpl.. & Inc.   & Cpl.  & Inc.  &  Cpl.  & Inc.\\ \midrule
$P$ 		& $0.93 \pm 3e^{-2}$  & $0.78 \pm 1e^{-2}$	& $0.93 \pm 3e^{-2}$ & $0.78 \pm 1e^{-2}$ & $0.97 \pm 2e^{-3}$  & $0.79 \pm 1e^{-3}$  & $0.93 \pm 4e^{-3}$   & $0.77 \pm 3e^{-3}$  & $0.97 \pm 2e^{-3}$     & $0.79 \pm 1e^{-3}$\\
$R$     		& $0.80 \pm 2e^{-2}$  & $0.92 \pm 3e^{-2}$ 	& $0.79 \pm 8e^{-3}$ 	& $0.92 \pm 4e^{-2}$ & $0.80 \pm 2e^{-2}$  & $0.97 \pm 2e^{-3}$  & $0.70 \pm 5e^{-3}$   & $0.92 \pm 5e^{-3}$ & $0.80 \pm 1e^{-3}$  & $0.97 \pm 2e^{-3}$ \\
$F1$   	& $0.86 \pm 8e^{-3}$ 	& $0.84 \pm 1e^{-2}$ 	& $0.86 \pm 1e^{-2}$ 	& $0.84 \pm 2e^{-2}$ & $\mathbf{0.88 \pm 1e^{-3}}$  & $\mathbf{0.87 \pm 1e^{-3}}$& $0.85 \pm 3e^{-3}$& $0.84 \pm 3e^{-3}$ 	& $\mathbf{0.88 \pm 2e^{-4}}$  & $\mathbf{0.87 \pm 5e^{-4}}$ \\ \midrule

$P$  & $0.91 \pm 2e^{-2}$ & $0.77 \pm 6e^{-3}$ & $0.95 \pm 9e^{-3}$ & $0.78 \pm 5e^{-3}$ & $0.73 \pm 1e^{-1}$  & $0.76 \pm 7e^{-2}$  & $0.95\pm 9e^{-3}$ & $0.67 \pm 1e^{-3}$ & $0.97 \pm8e^{-3}$  & $0.79\pm 3e^{-3}$\\
$R$     & $0.80 \pm 6e^{-3}$ &$0.90 \pm 3e^{-2}$ & $0.80 \pm 8e^{-3}$ & $0.95 \pm 1e^{-2}$ & $0.84\pm 1e^{-1}$  & $0.54 \pm 2e^{-2}$  & $0.64 \pm 5e^{-3}$  & $0.95\pm 8e^{-3}$ & $0.79 \pm 8e^{-3}$  & $0.97 \pm 4e^{-3}$ \\
$F1$   & $0.85 \pm 1e^{-2}$ &$0.83 \pm 1e^{-2}$ & $\mathbf{0.87 \pm 3e^{-3}}$ & $0.86 \pm 3e^{-3}$ & $0.76\pm3e^{-2}$  & $0.60\pm1e^{-1}$  & $0.76 \pm 1e^{-3}$ & $0.79\pm2e^{-3}$ & $\mathbf{0.87 \pm 8e^{-3}}$  & $\mathbf{0.87 \pm 3e^{-3}}$\\ \bottomrule
\end{tabular}
\end{adjustbox}
\end{table}



\section{Conclusion}
\label{sec:conclusion}

This work presents $\Delta$-MADS, a hybrid derivative-free optimization algorithm applied to solving the hyperparameter optimization problem of a variational autoencoder capable of adequately detecting anomalous data sent by the Mars Curiosity rover to the Mars science laboratory. The positive results obtained, especially on detecting anomalies, show the importance of such tools to assist the human experts in dealing with this type of unsupervised anomaly detection problems. The numerical results show that $\Delta$-MADS is able to score better than its two components: $\Delta$-DOGS and HyperNOMAD separately plus, it also allows to find better configurations with less computational budget compared to other schemes. 
Additionally, the algorithm can be used in broader applications since it does not rely on any prior knowledge on the dataset or any other bias. 

\section*{Broader Impact}
The authors believe that a broader impact discussion is not applicable.

\section*{Acknowledgments and Disclosure of Funding}
The research was carried out at the Jet Propulsion Laboratory, California Institute of Technology, under a contract with the National Aeronautics and Space Administration. It was supported in part by the MITACS Globalink program, the grant BFSD from the department of applied mathematics and industrial engineering of Polytechnique Montreal, the NSERC CRD RDCPJ 490744–15 grant and by an InnovÉE grant, both in collaboration with Hydro-Québec and Rio Tinto.
The authors would like to thank Brian Kahovec, Darisuh Divsalar, and David Hanks for the helpful discussions and support. 
\small
\bibliographystyle{plain}
\bibliography{bibliography.bib}


\end{document}